\documentclass[12pt]{article}
\usepackage{iclr2025_conference, times}

\usepackage{amsmath,amsfonts,bm}

\def\eqref#1{equation~\ref{#1}}

\def\1{\bm{1}}

\def\ve{{\bm{e}}}

\def\vt{{\bm{t}}}

\def\vv{{\bm{v}}}

\def\mA{{\bm{A}}}

\def\mC{{\bm{C}}}

\def\mE{{\bm{E}}}

\def\mH{{\bm{H}}}
\def\mI{{\bm{I}}}

\def\mM{{\bm{M}}}

\def\mP{{\bm{P}}}

\def\mS{{\bm{S}}}

\def\mY{{\bm{Y}}}
\def\mZ{{\bm{Z}}}

\DeclareMathAlphabet{\mathsfit}{\encodingdefault}{\sfdefault}{m}{sl}
\SetMathAlphabet{\mathsfit}{bold}{\encodingdefault}{\sfdefault}{bx}{n}

\def\sR{{\mathbb{R}}}

\usepackage{hyperref}
\usepackage{url}
\usepackage{setspace} 
\usepackage{booktabs}
\usepackage{graphicx}
\usepackage{multirow}

\usepackage[utf8]{inputenc}
\usepackage{setspace} 
\usepackage{geometry} 
\usepackage{times} 
\usepackage{enumitem}

\usepackage{dsfont}
\usepackage[table]{xcolor}
\usepackage{float}

\title{SLIP: Structural-aware Language-Image Pretraining for Vision-Language Alignment}

\author{Wenbo Lu
\thanks{ Under the guidance of Prof. Qiaoyu Tan}
\\ Department of Data Science\\ New York University Shanghai\\
\texttt{wenbo.lu@nyu.edu} \\ }

\newcommand{\best}[1]{\cellcolor[gray]{0.9}\textbf{#1}}

\iclrfinalcopy 

\begin{document}
    \maketitle

    \begin{abstract}
        Vision–Language Pretraining (VLP) has achieved remarkable success across
        various downstream tasks, but such gains are largely driven by scaling
        up on training data. Yet, literature methods treat image-text pairs as
        isolated training examples; this neglects the rich relational structure naturally
        present in many domains, such as e-commerce product co-purchase graphs
        and social recommendation networks. Inspired by neuroscientific evidence
        that human encodes knowledge as relationship cognitive maps, we
        introduce \textbf{S}tructure-aware \textbf{L}anguage-\textbf{I}mage
        \textbf{P}retraining \textbf{(SLIP)}. SLIP integrates a structural
        contrastive loss to align modalities while also modeling relationships
        between neighboring entities in a structured graph. To support this paradigm,
        we construct a large-scale \textbf{Amazon Product Co-purchase Multimodal
        Graph Dataset}, enabling structured cross-modality supervision at scale.
        Experiment results show that SLIP consistently outperforms CLIP on cross-modal
        retrieval and classification tasks in both zero-shot and few-shot settings,
        showing the value of relational supervision for cross-modal alignment.
    \end{abstract}

    \section{Introduction}

    Vision-language alignment has emerged as a key challenge in multimodal
    representation learning, with recent pretraining approaches achieving remarkable
    success by learning from web-scale data, driving progress in multimodal
    tasks such as image-text retrieval, visual question answering (VQA), and
    image captioning \citet{gan2022vision}. Ground-breaking work CLIP~\citep{radford2021learning}
    has shown that a simple contrastive objective can yield state-of-the-art representations
    when scaled to millions of noisy image-text pairs, and such large-scale
    training has thus become the paradigm for vision-language foundation models.
    However, these web-scale corpora are notoriously \emph{noisy}: captions can
    be generic, off-topic, or mismatched to the image. As a result, the
    performance is capped by the suboptimal source of supervision.

    Prior works addressing this challenge generally fall into two main
    categories: (1) exploring alternative forms of supervision within the data—for
    example, \cite{yao2021filip} introduces token-level alignment—and (2)
    refining existing supervising labels through techniques like knowledge
    distillation or bootstrapped relabeling, as in BLIP~\citep{li2022blip}. While
    these approaches enhance representation quality, they share a common
    limitation: they treat each image–text pair in isolation, overlooking the
    rich structural relationships among entities that are inherent in many real-world
    datasets.

    Many domains exhibit strong underlying structure. In e-commerce, products
    form co-purchase graphs-``customers who bought X also bought Y''-while in
    vision datasets, images are often connected via shared attributes or hierarchical
    semantics (e.g., ``lion'' and ``tiger'' as big cats). Similarly, in knowledge
    bases, entities are arranged into richly annotated relational graphs. Yet,
    existing VLP models ignore these structures and instead learn from
    contrasting loosely related individual pairs, losing consistency and semantic
    grounding. Leveraging structured relations could help capture latent
    connections and contextualize visual and textual concepts beyond what any
    single caption can provide.

    This idea finds support in cognitive neuroscience. Humans do not store knowledge
    in isolation; rather, they organize it into \emph{cognitive maps}-internal
    graph-like representations that encode relations among entities and
    facilitate flexible reasoning~\citep{behrens2018cognitive}. These maps
    enable us to generalize across experiences by exploiting structured
    relationships: when shown a laptop, for example, a shopper may automatically
    retrieve related items like a charger or a mouse. Representational distances
    in the brain are shaped not just by feature similarity but by relational
    proximity in learned conceptual spaces. Inspired by this, we posit that
    relational structure should likewise inform how machines align visual and
    textual modalities. \looseness=-1

    In this paper, we introduce \textbf{S}tructure-aware \textbf{L}anguage-\textbf{I}mage
    \textbf{P}retraining \textbf{(SLIP)}, a framework that augments vision-language
    contrastive learning with structural supervision. SLIP assumes that image-text
    pairs are nodes in a relational graph, where edges reflect semantic or contextual
    proximity-such as co-purchase, co-view, or hierarchical links. Specifically,
    Contextual signals are first aggregated using \textbf{modality-specific
    Graph Attention Network (GAT)} layers and then fused to form a unified node
    representation. This representation is used in a \textbf{structural
    contrastive loss}, which encourages embeddings of graph-neighboring nodes to
    be close in the space. This encourages cross-modal alignment for each pair, and
    coherent placement of related items, so that learned representations can
    reflect alignment and relational proximity.

    To support research in structure-aware multimodal learning, we curate and release
    the \textbf{Multimodal Amazon Product Co-purchase Graph Dataset}, a large-scale
    dataset of image-text pairs connected via real-world co-purchase edges. The
    dataset spans a wide range of product categories, with each product node annotated
    with both an image and a textual title or description. Edges represent
    purchase patterns mined from user behavior, offering weak but meaningful supervision.
    This dataset serves as a realistic and scalable benchmark for structured
    vision-language representation learning, where both multimodal data and relational
    graphs are available.

    \noindent
    \textbf{Contributions.} Our work makes the following key contributions:

    \begin{itemize}[topsep=0em, itemsep=0.5em, parsep=0em, partopsep=0em]
        \item We propose SLIP, a simple and scalable framework for incorporating
            structural supervision into contrastive vision-language pretraining.
            SLIP introduces a structural contrastive objective that regularizes image-text
            embeddings based on instance-level graph connectivity.

        \item We introduce the Amazon Product Co-purchase Multimodal Graph, a
            new large-scale dataset of image-text pairs linked by relational
            structure. It offers a challenging and realistic test bed for evaluation.

        \item We demonstrate through extensive experiments that SLIP achieves state-of-the-art
            results on retrieval and classification tasks in few-shot settings,
            outperforming standard CLIP baselines and providing empirical
            evidence for the value of structural context. \looseness=-1
    \end{itemize}

    \section{Related Work}

    \paragraph{Vision-Language Pre-training.}
    Vision-language pre-training learns strong generalization over various downstream
    tasks. Over the years, several model architectures have been proposed to support
    different types of applications. Dual-encoder architectures, such as CLIP~\citep{radford2021learning}
    and ALIGN~\citep{jia2021scaling}, encode images and texts separately and
    align them in a shared embedding space using contrastive learning. Fusion-encoder
    models like ViLBERT~\citep{tan2019lxmert} and UNITER~\citep{chen2020uniter} instead
    perform deep cross-modal interactions by jointly encoding vision and language
    inputs. Encoder-decoder architectures, including VL-T5~\citep{cho2021unifying},
    SimVLM~\citep{wang2021simvlm}, and BLIP~\citep{li2022blip}, follow a generative
    modeling approach that decodes text conditioned on visual features. More
    recently, unified transformer architectures such as PaLI~\citep{chen2022pali}
    and Uni-Perceiver v2~\citep{wang2022uniperceiver} aim to support a wide variety
    of vision and vision-language tasks using a single, flexible framework. Alongside
    these architectural advances, pre-training objectives have also converged on
    a few key formulations: image-text contrastive learning~\citep{radford2021learning, yao2022filip, li2022blip},
    image-text matching~\citep{li2021alignbeforefuse, wang2021oscar}, and masked
    modeling objectives applied to language or vision tokens~\citep{yu2022coca, wang2022uniperceiver}.
    However, these methods overlook the relational structure that exists among
    entities in many real-world domains, which SLIP explores.

    \paragraph{Scene-graph for object-level alignment.}
    Early work injected intra‑image structure by coupling object graphs with
    captioners or retrieval models. GCN-LSTM~\citep{DBLP:conf/eccv/YaoPLM18} and
    VSRN~\citep{DBLP:conf/iccv/LiYXZL19} propagated object–object interactions to
    improve description and matching accuracy. Follow‑ups for grounding and VQA~\citep{DBLP:conf/cvpr/YangLYD20, DBLP:conf/iccv/LiYXZL19}
    showed analogous gains for referring-expression grounding by jointly scene-graphing
    image regions and linguistic phrases. More recently, ERNIE‑ViL~\citep{DBLP:journals/corr/abs-2006-16934}
    and Structure-CLIP~\citep{huang2024structureclip} pushed this idea to the pre-training
    stage, either by multi‑task scene‑graph prediction or by scene‑graph‑driven
    hard negatives on top of CLIP. These models innovate by exploiting object interactions,
    yet they remain constrained to individual image-caption pairs and require
    costly scene-graph annotations. Instead, SLIP operates on the instance-level
    graphs defined over the dataset (e.g., co-purchase graphs), requiring no object
    detection or scene parsing. This preserves scalability and remains
    complementary and composable with their scene-level techniques.

    \paragraph{Knowledge-graph fusion for alignment.}
    Models such as MKVSE~\citep{DBLP:journals/tomccap/FengHP23} and VQA-GNN~\citep{DBLP:conf/iccv/0002YRWL23}
    enrich vision–language embeddings with commonsense or domain KGs, while
    GraphCLIP~\citep{DBLP:journals/kbs/ScaringiFVC25} applies a GNN over art-metadata
    graphs to align with images. These systems excel when factual or taxonomic
    context is critical, but they introduce dependency on curated KGs and often treat
    graphs as another modality and require separate modality-specific encoders.
    By contrast, our graph comes directly from user behavior and uses the same contrast heads 
    without extra encoders, making the technique broadly deployable even where external KGs are unavailable.

    \clearpage

    \section{Preliminaries}
    \label{sec:prelim}

    \subsection{Background on CLIP}
    \label{sec: clip background}

    \paragraph{Information Noise-contrastive Estimation (InfoNCE) loss.}
    \begin{figure}
        \centering
        \includegraphics[width=0.8\linewidth]{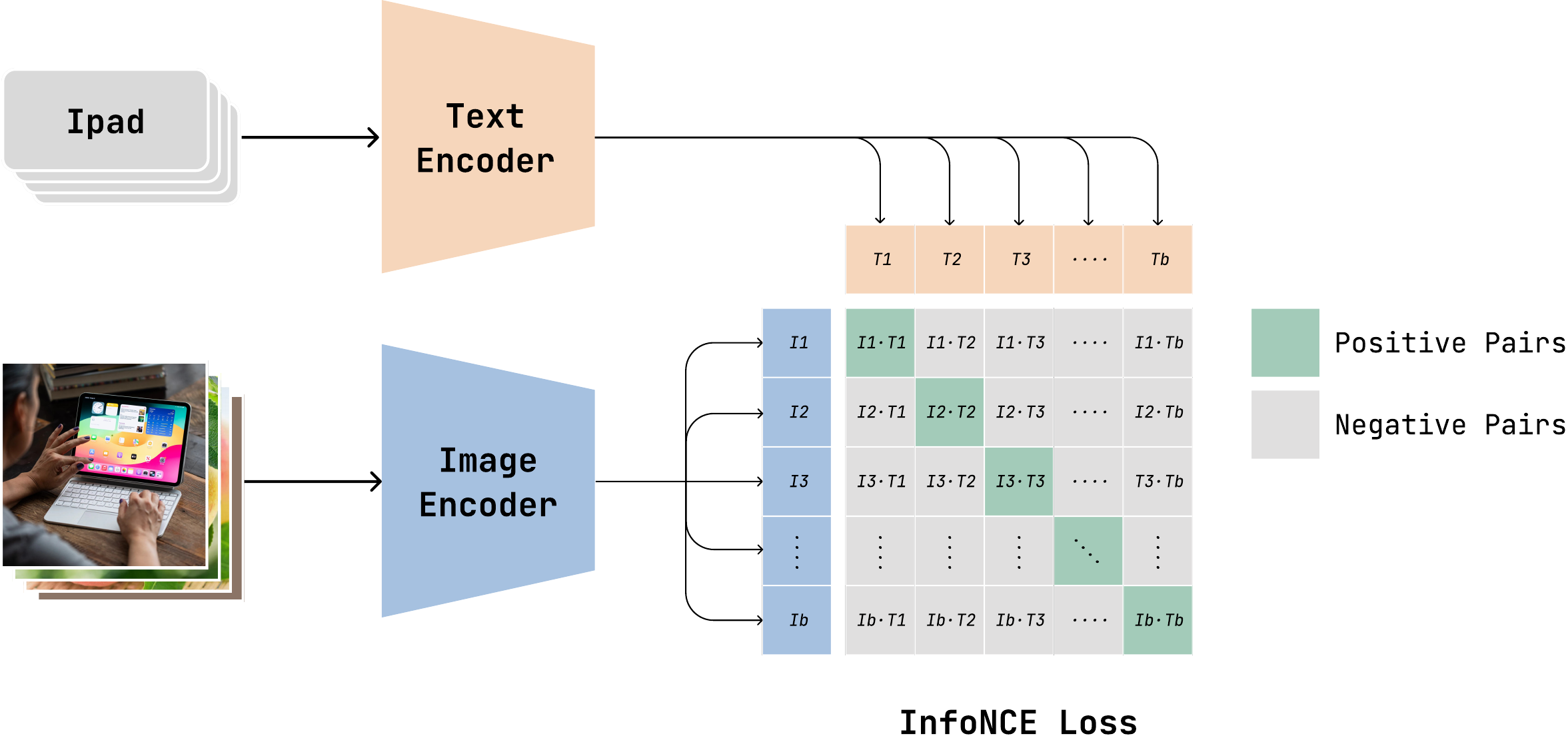}
        \caption{Illustration of contrastive vision-language pretraining using
        InfoNCE loss. An Amazon product (e.g., a laptop) is paired with its associated
        textual description to form positive image-text pairs. The model is
        trained to align these pairs while pushing apart negative pairs sampled
        from the batch. The image and text are independently encoded via modality-specific
        encoders before similarity computation.}
        \label{fig:contrastive learning}
    \end{figure}

    Given a set of $N$ image-text pairs $\{(\vv_{i}, \vt_{i})\}_{i=1}^{N}$ , the
    CLIP framework encodes each image $\vv_{i}$ using an image encoder $f_{V}$ and
    each text $\vt_{i}$ using a text encoder $f_{T}$ , producing $\ell_{2}$ -normalized
    embedding vectors $\ve_{v}^{i}= f_{V}(\vv_{i}) \in \sR^{d}$ and
    $\ve_{t}^{i}= f_{T}(\vt_{i}) \in \sR^{d}$ respectively. For a batch of size $b$
    , we denote the image and text embeddings as matrices $\mE_{v}\in \sR^{b
    \times d}$ and $\mE_{t}\in \sR^{b \times d}$ .

    The core learning objective in CLIP is a symmetric information noise-contrastive
    estimation (InfoNCE) loss~\citep{oord2018representation} that jointly aligns
    image-to-text and text-to-image representations. Specifically:
    \[
        \mathcal{L}_{v \rightarrow t}= -\frac{1}{N}\sum_{i=1}^{N}\log \frac{\exp\left(
        \ve_{v}^{i}\cdot \ve_{t}^{i}/ \tau \right)}{\sum_{j=1}^{N}\exp\left( \ve_{v}^{i}\cdot
        \ve_{t}^{j}/ \tau \right)}, \quad \mathcal{L}_{t \rightarrow v}= -\frac{1}{N}
        \sum_{j=1}^{N}\log \frac{\exp\left( \ve_{v}^{j}\cdot \ve_{t}^{j}/ \tau
        \right)}{\sum_{i=1}^{N}\exp\left( \ve_{v}^{i}\cdot \ve_{t}^{j}/ \tau \right)}
    \]
    where $\tau$ is a learnable temperature parameter. The numerator corresponds
    to the similarity between the $i$ -th image and its paired text, while the
    denominator sums over all possible text embeddings in the batch, normalizing
    the probabilities.

    In implementation, this is often computed as a symmetric cross-entropy loss
    using the similarity logits. As shown in Fig.~\ref{fig:contrastive
    learning}, the text and image logits matrices produced by the two encoders
    are multiplied to get the score for every pair efficiently (matrix multiplication
    is highly optimized on modern GPUs). Let $Y \in \sR^{b}$ denote the ground-truth
    labels (typically indices $\{0, 1, \dots, b-1\}$ ). The similarity logits are
    given by:
    \[
        \hat{\mY}_{v}= \exp(\tau) \mE_{v}\mE_{t}^{\top}, \qquad \hat{\mY}_{t}= \hat
        {\mY}_{v}^{\top},
    \]
    the numerator $\exp\left(\ve_{v}^{i}\cdot \ve_{t}^{i}/ \tau\right)$ becomes
    the $i$ -th diagonal element of $\hat{\mY}_{v}$ , and the denominator $\sum_{j=1}
    ^{N}\exp\left(\ve_{v}^{i}\cdot \ve_{t}^{j}/ \tau\right)$becomes the$i$-th row
    sum of $\hat{\mY}_{v}$ . This allows us to express the loss in terms of cross-entropy,
    where the ground-truth labels $Y$ are the indices of the diagonal elements. Recall
    that the general form of the cross-entropy loss for a predicted probability distribution
    $\hat{p}$ and a ground-truth one-hot label distribution $q$ is:
    \[
        \text{CE}(\hat{p}, q) = -\sum_{i}q_{i}\log \hat{p}_{i}.
    \]

    The contrastive loss used is the average of these two directional losses:
    \[
        \mathcal{L}_{\text{CLIP}}= \frac{1}{2}\left[ \mathcal{L}_{v \rightarrow
        t}+ \mathcal{L}_{t \rightarrow v}\right].
    \]

    Using this, the InfoNCE loss can be rewritten as a cross-entropy loss over
    the similarity logits $\hat{\mY}_{v}$ and the ground-truth labels $Y$ .
    Similarly, the text-to-image loss $\mathcal{L}_{t \rightarrow v}$ is derived
    in the same way. The final CLIP loss is computed as the mean of the cross-entropy
    losses from both modalities:
    \begin{equation}
        \mathcal{L}_{\text{CLIP}}= \frac{1}{2}\left[ \text{CE}\left(\hat{\mY}_{v}
        , Y\right) + \text{CE}\left(\hat{\mY}_{t}, Y\right) \right]. \label{eq: clip
        loss}
    \end{equation}

    \paragraph{Cross-Modal Alignment Score.}
    To evaluate how well a model aligns modalities, the average cosine
    similarity between matched image-text pairs can be a good measure. A higher alignment
    score indicates stronger semantic correspondence and a smaller modality gap between
    visual and textual embeddings. Perfect alignment is achieved when
    $\ve_{v}^{i}= \ve_{t}^{i}$ for all $i$, i.e., the paired image and text are
    mapped to the same point in the shared embedding space. Formally:
    \[
        \text{Alignment}= \frac{1}{N}\sum_{i=1}^{N}\ve_{v}^{i}\cdot \ve_{t}^{i},
        \qquad \text{Alignment}\in [-1, 1].
    \]

    \subsection{Multi-modal Graph Dataset}

    \paragraph{Multimodal Graph Dataset.}
    We assume a collection of $N$ multimodal items, each consisting of an \emph{image}
    and a textual \emph{description}. Formally, let:
    \[
        \mathcal{D}\;=\;\bigl\{(\vv_{i}, \vt_{i})\bigr\}_{i=1}^{N},
    \]
    where $v_{i}$ is an image and $t_{i}$ is the accompanying text (title or
    caption) of the $i$ -th item. Items are connected by a sparse, instance-level
    graph:
    \[
        \mathcal{G}\;=\; (\mathcal{V},\mathcal{E}), \qquad \mathcal{V}= \{1,\dots
        ,N\},
    \]
    where a node indices an image–text pair and an edge $(i,j)\in\mathcal{E}$ encodes
    semantic proximity (e.g., co-purchase, co-view, or knowledge-graph relation).
    We denote the binary adjacency matrix by $\mA\in\{0,1\}^{N\times N}$ and
    write We denote the binary adjacency matrix by $\mA\in\{0,1\}^{N\times N}$
    and write $\mA_{ij}=1$ if $(i,j)\in\mathcal{E}$ , $\mA_{ii}=0$ otherwise.

    \paragraph{Graph Masks.}
    For a user-specified hop threshold $h$ (default $h\!=\!1$ ), we construct a binary
    \emph{positive mask}
    \begin{equation}
        \mM_{ij}^{+}\;=\; \mathds{1}\bigl[\text{dist}_{\mathcal{G}_b}(i,j) \le h\bigr
        ], \qquad \mM_{ii}^{+}= 0, \label{eq: positive mask}
    \end{equation}
    and \emph{negative mask} $\mM^{-}= \mathds{1}- \mM^{+}- \mI_{b}$ .

    \section{Methodology}
    \label{sec:method}

    \begin{figure}[h]
        \centering
        \includegraphics[width=\linewidth]{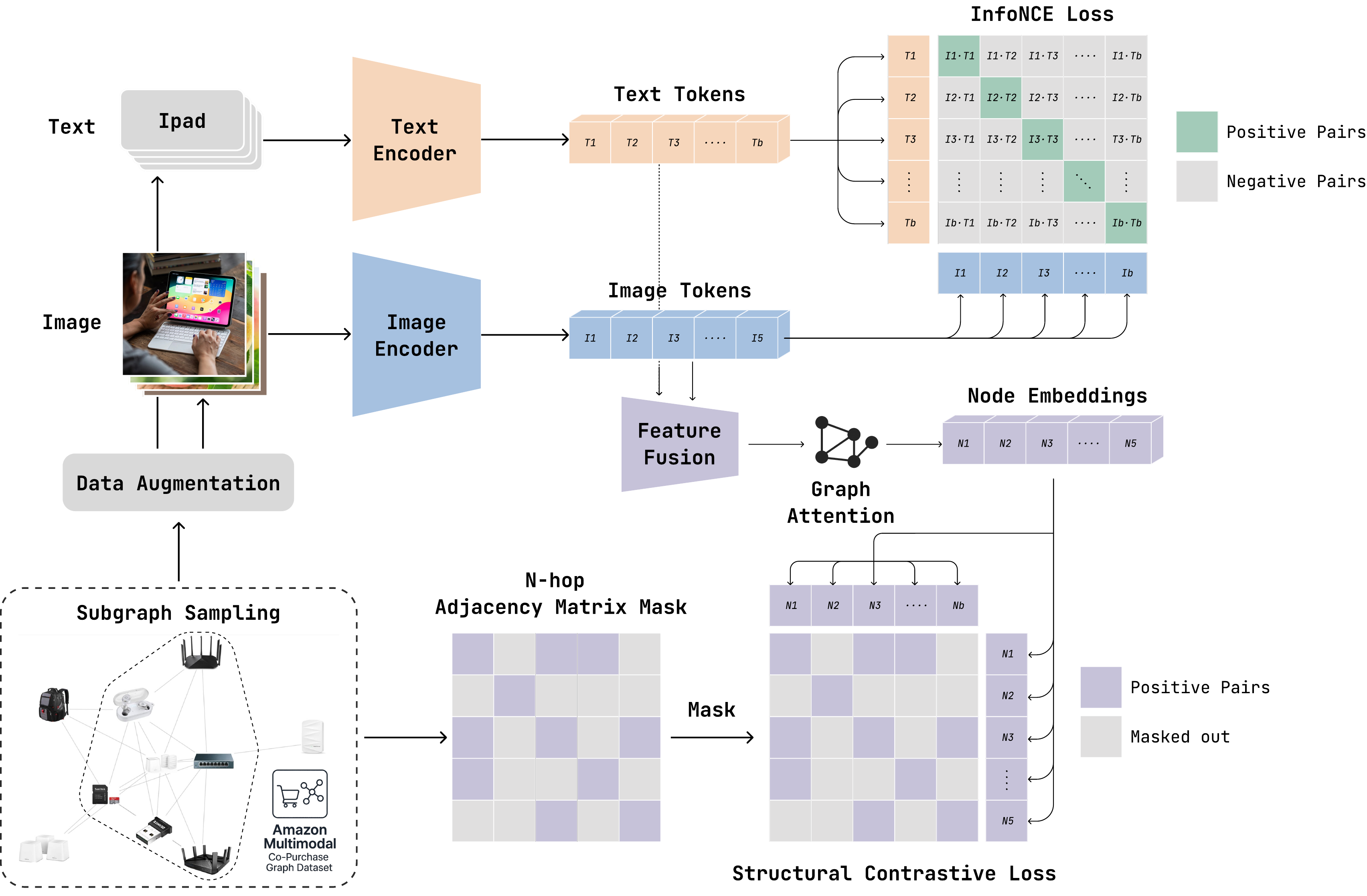}
        \caption{ \textbf{SLIP training pipeline.} A mini-batch of products is
        first sampled as a sub-graph from the Amazon co-purchase network (bottom
        left). Images undergo standard data augmentation, while titles/descriptions
        are tokenized. Both modalities are encoded by a CLIP backbone. \emph{Top
        path.} The image–text token similarities form the usual InfoNCE matrix:
        the diagonal (green) contains true pairs, off-diagonal cells (grey) act as
        negatives. \emph{Bottom path.} The sampled sub-graph is converteFd to an $n$
        -hop adjacency mask that selects nodes within one hop as additional
        positives (purple) and masks the rest (light grey). Image and text features
        are concatenated, passed through two layers of graph attention, and projected
        to node embeddings. Applying the mask to their similarity matrix yields
        the \emph{structural contrastive loss}.}

        \label{fig:overview}
    \end{figure}

    Fig.~\ref{fig:overview} illustrates an overview of the proposed Structure-aware
    Language–Image Pretraining (SLIP). Firstly, SLIP incorporates modality-specific
    Graph Attention Network (GAT) layers to encode structured relationships in images
    and texts, enhancing fine-grained relational representations with instance-level
    graphs (right part of Fig.~\ref{fig:overview}). Secondly, we introduce a structural
    contrastive loss that explicitly aligns the embeddings of structurally related
    nodes, integrating relational knowledge into the shared embedding space (right
    part of Fig.~\ref{fig:overview}). We will detail the design of the modality-specific
    GAT layers in Sec.~\ref{sec:architecture_design}, followed by the formulation
    of the structural contrastive loss in Sec.~\ref{sec:contrastive_learning_objective}.
    Additionally, we will introduce the curated Multimodal Amazon Product Co-purchase
    Graph Dataset in Sec.~\ref{sec:dataset}, which serves as a benchmark for
    evaluating the effectiveness of structure-aware vision-language pretraining
    methods like SLIP.

    \subsection{Architecture Design}
    \label{sec:architecture_design}

    We start from pretrained CLIP dual encoders: an image encoder $f_{V}$ and a
    text encoder $f_{T}$ , which produce normalized embeddings
    $E_{v}, E_{t}\in \mathbb{R}^{b\times d}$ from a batch of $b$ image-text
    pairs. While these encoders effectively capture cross-modal correlations,
    they do not consider relationships among distinct image-text pairs. To address
    this limitation, we integrate relational information from an instance graph
    $\mathcal{G}_{b}$ associated with the current mini-batch. Each node in $\mathcal{G}
    _{b}$ corresponds to an image-text pair, with edges indicating semantic
    proximity (e.g., co-purchase relationships). To propagate relational information,
    we apply two layers of Graph Attention Networks (GAT)~\citep{velivckovic2017graph}
    separately on the visual and textual embeddings:
    \begin{equation}
        \mH_{V}^{(l+1)}= \text{GAT}(\mH_{V}^{(l)}, \mA), \quad \mH_{T}^{(l+1)}= \text{GAT}
        (\mH_{T}^{(l)}, \mA), \quad l = 0,1,
    \end{equation}
    where $H_{V}^{(0)}=E_{v}$ , $H_{T}^{(0)}=E_{t}$ , and
    $\mA \in \{0,1\}^{b \times b}$ denotes the adjacency matrix. Through attention-based
    aggregation, $H_{V}^{(2)}$ and $H_{T}^{(2)}$ encode neighborhood context
    specific to their modalities, reflecting visual similarity or textual relatedness.
    Then to integrate these modality-specific relational embeddings, we concatenate
    and project them into a common node embedding space via a lightweight projection:
    \begin{equation}
        \mZ = \phi(\left[\mH_{V}^{(2)}\,\|\,\mH_{T}^{(2)}\right]) \in \mathbb{R}^{b\times
        d},
    \end{equation}
    where $\phi$ is a single projection with nonlinear activation and
    normalization, and $\mZ$ is $\ell_{2}$ -normalized. This node embedding $Z$ explicitly
    encodes structural relationships between nodes and serves as the basis for structural
    contrastive learning.

    \subsection{Structural Contrastive Learning Objective}
    \label{sec:contrastive_learning_objective}

    We design a structural contrastive loss, an adaptation of the InfoNCE loss defined
    in Eq.~\ref{eq: clip loss}, to explicitly inject structural supervision. In
    the original InfoNCE (used in CLIP), each sample has exactly one positive (the
    paired text or image), and the rest of the batch serves as negatives. The model
    learns to maximize similarity with the positive and minimize it with the negatives
    using a softmax-based cross-entropy formulation.

    We generalize this idea to graph settings: instead of treating only the
    matched pair as the positive, we treat all graph-connected nodes as positives.
    That is, the positive set becomes a masked subset of the batch defined by the
    graph structure. This generalization requires modifying the loss to handle
    multiple positives and an arbitrary mask of negatives. We formulate as follows:
    let $\mZ \in \sR^{b \times d}$ be the matrix of normalized node embeddings
    output by GAT layers, representing a batch of $b$ multimodal items. We
    define the pairwise similarity matrix as
    $\mS= \tau^{-1}\cdot \mZ \cdot \mZ^{\top}$ , where $\tau$ is a temperature
    parameter. In order to turn the similarity scores into probabilities over the
    batch, we apply a row-wise softmax operation. Notably, we use log-softmax
    similarly to the original InfoNCE to avoid numerical instability:
    \[
        \log \mP_{i,j}= \log \frac{\exp(S_{ij})}{\sum_{k=1}^{b}\exp(S_{ik})}.
    \]

    The graph loss is given by:
    \begin{equation}
        \mathcal{L}_{\text{graph}}= -\frac{1}{\|\mM^{+}\| + \epsilon}\sum_{i,j}\mM
        ^{+}_{i,j}\cdot \log \mP_{i,j}, \label{eq: graph loss}
    \end{equation}
    where $\mM$ stands for the binary positive graph mask introduced in Eq.~\ref{eq:
    positive mask}, and $\|\mM^{+}\|$ denotes the number of positive pairs.

    To maintain the cross-modal alignment capability of CLIP, we retain the
    original implementation of InfoNCE loss as defined in Eq.~\ref{eq: clip loss},
    ensuring direct image–text correspondence. This loss encourages the model to
    align paired image and text embeddings while separating unpaired ones in the
    shared embedding space:
    \[
        \mathcal{L}_{\text{clip}}= \frac{1}{2}\left[ \text{CE}\left(\hat{\mY}_{v}
        , Y\right) + \text{CE}\left(\hat{\mY}_{t}, Y\right) \right].
    \]

    For tasks that benefit from class supervision, we optionally introduce an
    auxiliary classifier applied to the learned node features. Depending on whether
    the graph-enhanced path is active, the input features are either the GNN-fused
    representations or the concatenated outputs of the CLIP encoders. The classifier
    consists of a single linear projection $\mathbf{C}= \text{Linear}(\mZ)$ , where
    $\mZ\in \sR^{N \times d'}$ denotes the input features and $d'$ is the feature
    dimension (either $d$ or $2d$ , depending on fusion strategy). The output
    $\mathbf{C}\in \mathbb{R}^{N \times C}$ represents class logits for $C$
    target categories. To train the classification head, we use a standard cross-entropy
    loss over the predicted class distribution. Let $\mY_{i,c}\in \{ 0,1\}$ be the
    one-hot ground-truth label for instance $i$ and class $c$ , and $\text{softmax}
    (\mathbf{C}_{i})_{c}$ be the predicted probability. The loss is computed as:
    \begin{equation}
        \mathcal{L}_{\text{aux}}= -\frac{1}{N}\sum_{i=1}^{N}\sum_{c=1}^{C}\mY_{i,c}
        \cdot \log\left( \text{softmax}(\mC_{i})_{c}\right). \label{eq: aux loss}
    \end{equation}

    Therefore, the final training loss combines cross-modal alignment Eq.~\ref{eq:
    clip loss}, structural alignment Eq.~\ref{eq: graph loss}, and auxiliary classification
    loss Eq.~\ref{eq: aux loss}:
    \begin{equation}
        \mathcal{L}_{\text{total}}= \mathcal{L}_{\text{clip}}+ \lambda_{\text{graph}}
        \mathcal{L}_{\text{graph}}+ \lambda_{\text{aux}}\mathcal{L}_{\text{aux}},
    \end{equation}
    where we empirically pick hyperparameters $\lambda_{\text{graph}}= 0.05$ and
    $\lambda_{\text{aux}}= 0.1$ are set to balance the magnitude of instance-level
    modality alignment loss, structured relational coherence loss, and auxiliary
    classification supervision loss.

    \section{Dataset}
    \label{sec:dataset}

    Based on the Amazon Products dataset introduced by \cite{hou2024bridging}, we
    construct \textbf{Multimodal Amazon Product Co-purchase Graph Dataset}, ƒa comprehensive
    multimodal graph that integrates textual, visual, and structural information.
    Each product is represented through textual descriptions (titles and detailed
    specifications) and corresponding high-resolution product images. All products
    are organized within a hierarchical category taxonomy (e.g., ``Electronics
    $>$ Smartphones $>$ Accessories''). We select different category
    granularities across subsets based on class diversity and representativeness
    at that specific level.

    Purchase records initially create a bipartite graph between consumers and
    products. We derive our co-purchase graph by connecting products that share common
    purchasers (second-order connections). To ensure data quality and
    statistical robustness, we employ two filtering mechanisms: (1) k-core
    decomposition with $k=5$ , which recursively removes nodes with fewer than 5
    connections until all remaining nodes have at least 5 connections, preserving
    only the dense, stable subgraph where meaningful patterns can emerge; and (2)
    co-purchase frequency filtering, retaining only edges representing products
    co-purchased at least 3 times by different users. Frequency filtering is important
    for identifying meaningful product associations that represent actual shopping
    patterns rather than random coincidences. This dual filtering approach
    eliminates noise from sparse interactions, reduces the impact of outliers,
    and ensures that captured co-purchase relationships reflect genuine consumer
    behaviors rather than coincidental or one-time purchasing decisions.

    \begin{figure}
        \centering
        \includegraphics[width=0.5\linewidth]{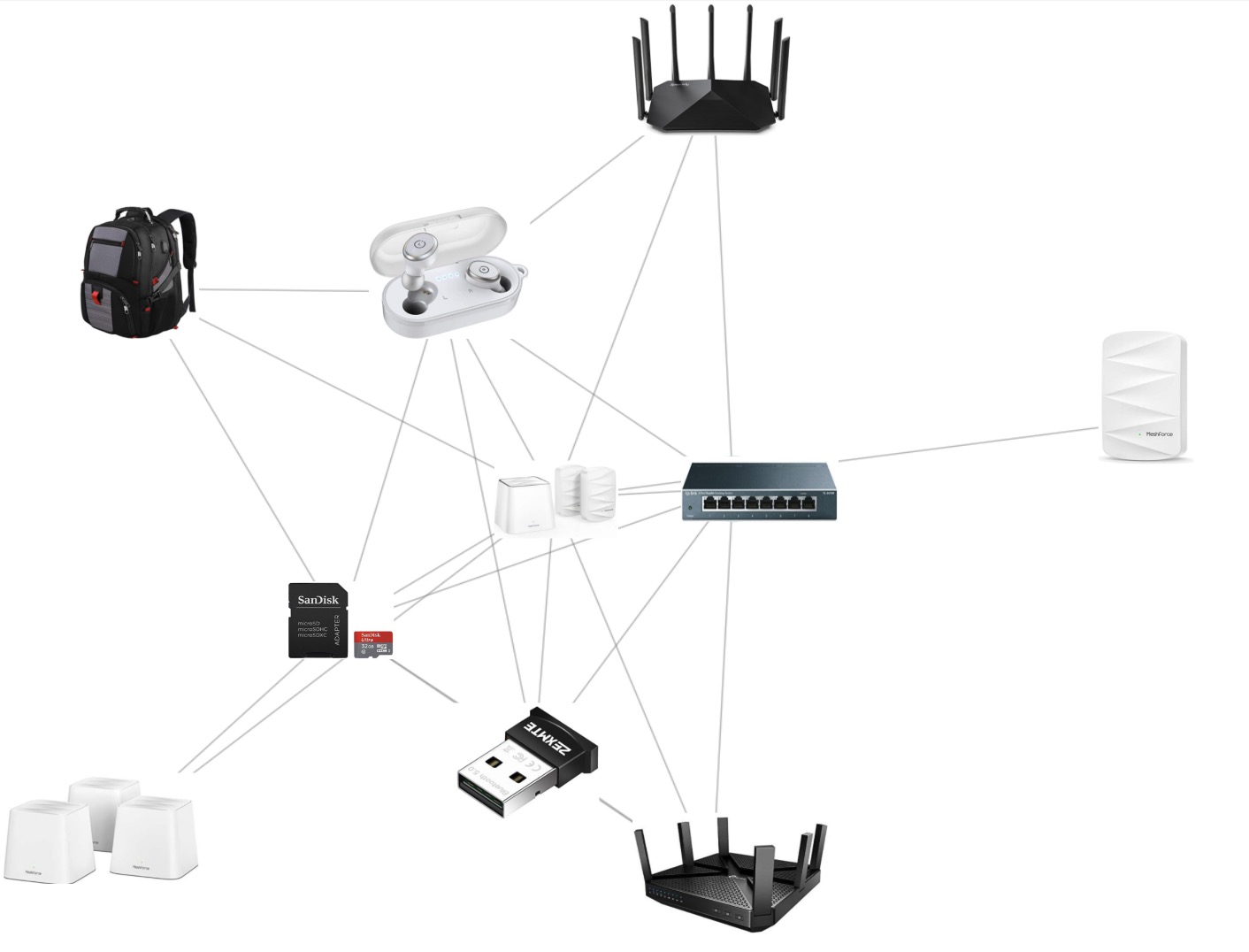}
        \caption{A 10-node subgraph sampled from the curated dataset (\texttt{Electronics})}
        \label{fig:sampled graph}
    \end{figure}

    Fig.~\ref{fig:sampled graph} illustrates an example product co-purchase graph
    segment, visually highlighting the structural connections among products
    such as routers, switches, and other electronic devices. Nodes indicate individual
    products, and edges represent co-purchase relationships. The graph structure
    reveals an intuitive pattern: immediate neighbors (first-hop connections)
    typically represent complementary products rather than similar ones, as consumers
    rarely purchase identical items multiple times, but instead buy components
    that work together (e.g., a laptop and its compatible charger, instead of
    buying two laptops). In contrast, second-hop neighbors—products connected through
    an intermediary node—tends to be more similar to the original item. This
    pattern emerges naturally from consumer behavior, where complementary purchases
    create first-hop connections, while similar products become linked
    indirectly through their shared complementary items, forming a cohesive network
    of related products with distinct first-hop and second-hop relationship characteristics.

    \begin{table}[hbt]
        \centering
        \begin{tabular}{l|r|r|c|c}
            \toprule \multirow{2}{*}{\textbf{Main Category}} & \multirow{2}{*}{\textbf{Nodes (k)}} & \multirow{2}{*}{\textbf{Edges (k)}} & \multicolumn{2}{c}{\textbf{CLIP-T Score $\pm \text{ Std.dev}$ }} \\
            \cmidrule{4-5}                                   &                                     &                                     & \textbf{Title}                                                  & \textbf{Description} \\
            \midrule \midrule Video Games                    & 13                                  & 233                                 & 33.9 $\pm$ 4.3                                                  & 30.6 $\pm$ 4.5       \\
            Baby Products                                    & 14                                  & 272                                 & 31.7 $\pm$ 4.1                                                  & 31.0 $\pm$ 4.0       \\
            Office Products                                  & 16                                  & 90                                  & 32.8 $\pm$ 4.4                                                  & 30.5 $\pm$ 4.5       \\
            Arts/Crafts/Sewing                               & 28                                  & 197                                 & 31.8 $\pm$ 4.6                                                  & 29.4 $\pm$ 4.8       \\
            CDs \& Vinyl                                     & 36                                  & 845                                 & 26.3 $\pm$ 6.1                                                  & 29.1 $\pm$ 5.9       \\
            Grocery \& Food                                  & 50                                  & 1,042                               & 35.5 $\pm$ 4.3                                                  & 32.4 $\pm$ 4.9       \\
            Automotive                                       & 57                                  & 273                                 & 30.8 $\pm$ 3.8                                                  & 28.8 $\pm$ 3.9       \\
            Toys \& Games                                    & 58                                  & 395                                 & 33.3 $\pm$ 3.9                                                  & 31.2 $\pm$ 4.5       \\
            Movies/TV                                        & 60                                  & 2,118                               & 33.7 $\pm$ 5.1                                                  & 31.1 $\pm$ 4.9       \\
            Health/Household                                 & 73                                  & 1,436                               & 34.1 $\pm$ 4.6                                                  & 32.3 $\pm$ 4.9       \\
            Beauty \& Care                                   & 87                                  & 1,841                               & 32.8 $\pm$ 4.1                                                  & 30.9 $\pm$ 4.3       \\
            Electronics                                      & 98                                  & 2,015                               & 31.4 $\pm$ 3.7                                                  & 29.3 $\pm$ 3.9       \\
            Clothing/Shoes                                   & 172                                 & 1,873                               & 31.3 $\pm$ 3.4                                                  & 29.5 $\pm$ 3.9       \\
            Books                                            & 194                                 & 3,988                               & 32.9 $\pm$ 5.0                                                  & 28.9 $\pm$ 4.7       \\
            \bottomrule
        \end{tabular}
        \caption{Multimodal Amazon Product Co-purchase Graph Dataset statistics
        with CLIP-T scores (mean $\pm$ std), evaluating the semantic alignment between
        textual descriptions of products and their images. Nodes and edges are
        counted in thousands (k), excluding entries with missing or non-compliant
        data (invalid images, empty titles/descriptions). CLIP scores quantify text-image
        alignment to filter low-quality pairs.}
        \label{tab:dataset stats}
    \end{table}

    Tab.~\ref{tab:dataset stats} summarizes key statistical properties of the
    processed Amazon product dataset after rigorous data quality checks. To ensure
    robust multimodal alignment, we filtered out entries with incomplete data, such
    as missing titles, inadequate descriptions, or low-quality images. Semantic
    alignment between textual and visual components is quantified using CLIP-T
    scores, providing a measure of the coherence between images and their corresponding
    textual descriptions. Higher scores reflect stronger semantic congruence, guiding
    the systematic exclusion of ambiguous or mismatched product representations that
    could degrade model training and evaluation.

    Overall, the Amazon Products Multimodal Graph presents an ideal and robust
    benchmark for evaluating structure-aware multimodal models~\citep{lu2022survey}.
    By combining text, image, and structured contextual information, this dataset
    enables a comprehensive assessment of models' capabilities to effectively
    integrate multimodal inputs and exploit relational contexts to enhance
    performance on tasks such as product retrieval, classification, and
    recommendation.

    \section{Experiments}

    Our experimental evaluation aims to assess the effectiveness of SLIP against
    existing state-of-the-art contrastive vision-language models. We first
    describe the experimental setup (Sec.\ref{sec:exp_setup}), then present results
    on retrieval tasks (Sec.~\ref{sec:retrieval_results}), followed by
    ablation studies exploring key design choices (Sec.~\ref{sec:ablations}).

    \subsection{Experimental Setup}
    \label{sec:exp_setup}

    \paragraph{Implementation.}
    We implement our structure-aware approach by extending the CLIP architecture
    \citep{radford2021learning}. Specifically, we utilize the \texttt{openai/clip-vit-base-patch16
    (patch32)} variants, which employs a ViT-B/16 (32) vision encoder and a transformer-based
    text encoder. Our graph-aware components consist of dual Graph Attention
    Networks (GATs) that process image and text embeddings separately before
    fusion. Each GAT layer uses a hidden dimension of 512 units with 4 attention
    heads and a dropout rate of 0.1 for regularization. This configuration
    allows the model to capture and propagate structural information across the
    graph while maintaining computational efficiency.

    \paragraph{Training Details.}
    A series of optimization strategies is adopted to enhance training efficiency
    and model performance. Our approach employs discriminative fine-tuning with
    varying learning rates across the model hierarchy, where deeper layers use the
    base rate of $1\times10^{-5}$ while gradually increasing rates for shallower
    layers by a factor of 0.8. This approach allows different layers to adapt at
    appropriate rates. Shallower layers receive higher rates to adjust low-level
    feature extraction, while deeper layers handling semantic information
    receive lower rates for more subtle refinement. This optimization strategy improves
    overall model performance by recognizing the distinct roles of different network
    components. The graph components are optimized separately with a higher
    learning rate ( $4 \times 10^{-3}$ ) than the pre-trained CLIP components to
    allow faster adaptation to the structural information as we are training
    from scratch. The training schedule extends up to 50 epochs with early stopping
    based on the validation metric, with a patience of 10 epochs and a minimum
    delta of 0.001. To manage memory constraints while training with larger batch
    sizes, we employ gradient checkpointing, trading computation for memory efficiency.
    We also implement a linear learning rate scheduler with warmup, using 500 warmup
    steps and gradually decreasing the learning rate to 0 for the remaining training
    steps to stabilize the early phases of training and prevent unstable gradients.

    \paragraph{Dataset Preparation.}
    For our experiments, we mainly focus on the \texttt{Electronics} subset of
    our curated Multimodal Amazon Product Co-purchase Graph Dataset introduced
    in Sec.~\ref{sec:dataset} for its moderate size. The dataset is split using
    a 60\%-10\%-30\% ratio for training, validation, and testing, respectively,
    ensuring sufficient data for model training while reserving a substantial
    portion for comprehensive evaluation. Through preliminary experiments, we
    determined that product titles serve as more informative textual components
    than full descriptions for our retrieval tasks, likely due to their concise nature
    and higher information density. Consequently, we primarily use product titles
    as the text modality in our main experiments.

    \paragraph{Evaluation Metrics.}
    We evaluate the performance of our model and baselines using standard
    information retrieval metrics that capture different aspects of retrieval
    quality. Mean Reciprocal Rank (MRR) serves as our primary metric, offering a
    balanced assessment of overall retrieval performance with emphasis on higher
    ranks. We also report Recall@K for K=1, 5, and 10 to measure the model's ability
    to retrieve relevant items within the top-K results. Additionally, we track Mean
    and Median Rank metrics to provide an overview of the distribution of retrieval
    positions across the test set.

    \paragraph{Baselines.}
    For our evaluation, we compare SLIP exclusively against the original CLIP model.
    We focus on this single baseline for several reasons: First, CLIP represents
    the canonical dual-encoder architecture for vision-language alignment and
    serves as the direct foundation for our work, making it the most relevant point
    of comparison. Second, CLIP's performance characteristics are well-documented
    and understood across the research community, providing a reliable benchmark.
    Third, our primary contribution is the integration of structural information
    into the contrastive learning framework, rather than architectural innovations
    that would necessitate comparison against the full spectrum of vision-language
    models. Finally, this focused comparison allows us to isolate the specific impact
    of our structural enhancements while controlling for other variables. By
    demonstrating improvements over CLIP, we establish a clear proof of concept for
    structure-aware vision-language alignment that can potentially extend to other
    architectures in future work.

    \subsection{Retrieval Results}
    \label{sec:retrieval_results}

    \begin{table}[h]
        \centering
        \begin{tabular}{@{}l|ccc|cc|ccc@{}}
            \toprule \multirow{2}{*}{Method} & \multicolumn{3}{c|}{\textbf{MRR} $\uparrow$ } & \multicolumn{2}{c|}{\textbf{Rank} $\downarrow$ } & \multicolumn{3}{c}{\textbf{Top- $k$ Recall} $\uparrow$ } \\
            \cmidrule{2-9}                   & i2t                                           & t2i                                              & Mean                                                    & Med      & Mean        & @1           & @5           & @10          \\
            \midrule CLIP (fine-tuned)       & 0.518                                         & 0.522                                            & 0.520                                                   & 2        & 133.4       & 0.403        & 0.659        & 0.743        \\
            SLIP (ours, graph)               & \best{0.585}                                  & \best{0.582}                                     & \best{0.584}                                            & \best{2} & \best{93.4} & \best{0.478} & \best{0.712} & \best{0.786} \\
            \bottomrule
        \end{tabular}
        \caption{Main comparison of cross-modal retrieval performance between
        our proposed SLIP model and baseline CLIP fine-tuning. We evaluate retrieval
        in both directions—Image-to-Text (I2T) and Text-to-Image (T2I)—using
        standard metrics: MRR, median and mean rank (lower is better), and top-
        $k$ recall at $k=1$ , $5$ , and $10$ (higher is better). All models are trained
        under the same optimization schedule. Best performance in each metric is
        highlighted.}
        \label{tab:main}
    \end{table}

    Our main results, presented in Tab.~\ref{tab:main}, demonstrate that SLIP consistently
    outperforms the fine-tuned CLIP baseline across all retrieval metrics. On
    the \texttt{Electronics} dataset, SLIP achieves an average MRR of 0.584
    compared to CLIP's 0.520, representing a substantial 12.3\% relative
    improvement. This performance gain is consistent across both retrieval directions
    (image-to-text and text-to-image), indicating that the structural
    information benefits bidirectional cross-modal alignment equally.

    The improvement in top-1 recall is particularly noteworthy, with SLIP achieving
    0.478 versus CLIP's 0.403, representing an 18.6\% relative gain. This
    suggests that structural context information significantly enhances the
    model's ability to identify the correct match as the top result. The improvements
    in recall at higher ranks (R@5, R@10) are also substantial but less dramatic,
    indicating that the benefits of structural context are most pronounced when discriminating
    between closely related candidates.

    Mean rank statistics further highlight SLIP's advantage, with our model
    achieving a mean rank of 93.4 compared to CLIP's 133.4. This 30\%
    improvement suggests that SLIP makes fewer catastrophic errors where the
    correct match is ranked very low. Interestingly, the median rank remains at
    2 for both models, indicating that both perform well on typical cases, but
    SLIP handles challenging examples more robustly.

    \begin{figure}[h]
        \centering
        \includegraphics[width=0.9\linewidth]{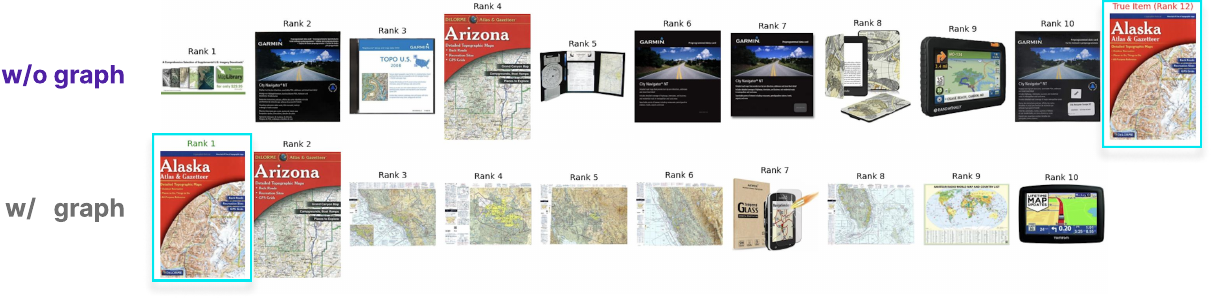}
        \caption{ Qualitative retrieval comparison for the query title \emph{``Garfrfin
        Delorme Atlas \& Gazetteer Paper Maps - Alaska, AA-000004-000''}. The top
        row shows the top ten image results from CLIP fine-tuned without graph supervision
        (w/o graph), and the bottom row shows the corresponding results from
        SLIP with graph supervision (w/ graph). True matches are highlighted
        with a colored border and annotated with their retrieval rank.}
        \label{fig:map_retrieval}
    \end{figure}
    Here we provide a visual example to better showcase SLIP's unique advantages.
    Fig.\ref{fig:map_retrieval} demonstrates the qualitative improvement in text-to-image
    retrieval when incorporating graph-aware supervision. It compares retrieval
    results for a specific product query about an Alaska atlas map. Without graph
    supervision, the fine-tuned CLIP model prioritizes visually similar
    items—such as generic road-trip photos and GPS device screenshots—placing
    the actual atlas at rank 12. This indicates a reliance on low-level visual
    cues rather than semantic context. In contrast, SLIP's graph-aware training
    improves relevance: the Alaska atlas appears at rank 1, followed by other
    region-specific maps (Arizona, detailed U.S. topographic sheets) and
    relevant cartographic products in the top 5. By injecting co-purchase and
    categorical relationships, SLIP learns to group semantically related map products
    together, yielding far more precise retrieval for specialized queries.

    \subsection{Ablation Studies}
    \label{sec:ablations}

    To understand the contribution of different components and hyperparameters in
    our framework, we conducted extensive ablation studies.

    \begin{table}[h]
        \centering
        \begin{tabular}{rr|ccc|cc|ccc}
            \toprule \multirow{2}{*}{Batch} & \multirow{2}{*}{Graph} & \multicolumn{3}{c|}{\textbf{MRR} $\uparrow$ } & \multicolumn{2}{c|}{\textbf{Rank} $\downarrow$ } & \multicolumn{3}{c}{\textbf{Top- $k$ Recall} $\uparrow$ } \\
            \cmidrule{3-10}                 &                        & i2t                                           & t2i                                              & Mean                                                    & Med      & Mean        & @1           & @5           & @10          \\
            \midrule 64                     & w/o G                  & 0.493                                         & 0.489                                            & \best{0.491}                                            & 3        & 117.3       & 0.367        & 0.638        & 0.729        \\
            64                              & w/ G                   & 0.434                                         & 0.430                                            & 0.432                                                   & 4        & 131.5       & 0.304        & 0.584        & 0.689        \\
            \midrule 128                    & w/o G                  & 0.503                                         & 0.502                                            & \best{0.502}                                            & 2        & 109.3       & 0.378        & 0.651        & 0.741        \\
            128                             & w/ G                   & 0.468                                         & 0.471                                            & 0.470                                                   & 3        & 135.2       & 0.346        & 0.618        & 0.711        \\
            \midrule 256                    & w/o G                  & 0.511                                         & 0.516                                            & 0.514                                                   & 2        & 134.3       & 0.395        & 0.656        & 0.742        \\
            256                             & w/ G                   & 0.512                                         & 0.518                                            & \best{0.515}                                            & 2        & 129.4       & 0.399        & 0.654        & 0.737        \\
            \midrule 512                    & w/o G                  & 0.519                                         & 0.523                                            & 0.521                                                   & 2        & 128.3       & 0.401        & 0.666        & 0.746        \\
            512                             & w/ G                   & 0.541                                         & 0.541                                            & \best{0.541}                                            & 2        & 118.5       & 0.429        & 0.678        & 0.757        \\
            \midrule 1,024                  & w/o G                  & 0.518                                         & 0.522                                            & 0.520                                                   & 2        & 133.4       & 0.403        & 0.659        & 0.743        \\
            1,024                           & w/ G                   & \best{0.585}                                  & \best{0.582}                                     & \best{0.584}                                            & \best{2} & \best{93.4} & \best{0.478} & \best{0.712} & \best{0.786} \\
            \bottomrule
        \end{tabular}
        \caption{Ablation study on the sub-graph batch size used during training.
        For each size, we compare a baseline CLIP fine-tune \emph{without} graph
        supervision (\textbf{w/o G}) against our structure-aware variant \emph{with}
        graph supervision (\textbf{w/ G}). Higher is better for MRR and Recall,
        lower is better for Rank. The best mean-MRR at each batch size is highlighted.}
        \label{tab:batch}
    \end{table}

    \paragraph{Impact of Batch Size.}
    Tab.~\ref{tab:batch} explores how graph-based learning interacts with batch
    size. For smaller batches (64 and 128), the baseline CLIP outperforms the
    graph-enhanced model. As batch size increases to 256, performance equalizes,
    and at larger batch sizes (512 and 1024), the graph-enhanced model
    demonstrates clear superiority. This pattern reveals a crucial insight:
    effective graph-based learning requires sufficient context to extract
    meaningful structural patterns. With small batches, the subgraphs are too
    sparse and disconnected to provide useful signals, potentially introducing noise
    instead. At 1024 batch size, where we observe maximal gain, the subgraph
    likely contains enough connected components and structural diversity to
    enable effective propagation of contextual information.

    \begin{table}[htbp]
        \centering
        \begin{tabular}{@{}lccc|ccc|cc|ccc@{}}
            \toprule \multicolumn{4}{c|}{Configuration} & \multicolumn{3}{c|}{\textbf{MRR} $\uparrow$ } & \multicolumn{2}{c|}{\textbf{Rank} $\downarrow$ } & \multicolumn{3}{c}{\textbf{Top- $k$ Recall} $\uparrow$ } \\
            \midrule \textbf{Variant}                   & G                                             & Aux                                              & DLR                                                     & i2t          & t2i          & Mean         & Med      & Mean        & @1    & @5    & @10          \\
            \midrule \midrule CLIP fine-tune (baseline) & -                                             & -                                                & \checkmark                                              & 0.518        & 0.522        & 0.520        & 2        & 133.4       & 0.403 & 0.659 & 0.743        \\
            \quad +\,Graph only                         & \checkmark                                    & -                                                & \checkmark                                              & \best{0.597} & \best{0.596} & \best{0.597} & 2        & 114.1       & 0.492 & 0.725 & 0.790        \\
            \quad +\,Graph\,+\,Aux                      & \checkmark                                    & \checkmark                                       & \checkmark                                              & 0.585        & 0.582        & 0.584        & \best{2} & \best{93.4} & 0.478 & 0.712 & \best{0.786} \\
            \quad +\,Graph\,+\,Aux, no DLR              & \checkmark                                    & \checkmark                                       & -                                                       & 0.565        & 0.567        & 0.566        & 2        & 134.4       & 0.460 & 0.695 & 0.764        \\
            \bottomrule
        \end{tabular}
        \caption{Component ablation on the large‐batch (1,024) setting. All models
        use the same base learning rate $1.6\times10^{-4}$ . We successively add
        graph supervision (G), the auxiliary classification head (Aux), and the discriminative
        layer-wise learning-rate schedule (DLR) for each layer decay by $0.8$ . Higher
        is better for MRR and Recall; lower is better for Rank.}
        \label{tab:ablation}
    \end{table}

    \paragraph{Component Analysis.}
    Tab.~\ref{tab:ablation} isolates the contribution of individual components
    in our framework. The "Graph only" variant, which incorporates structural
    supervision without auxiliary classification, achieves the highest MRR (0.597).
    This suggests that graph-based contrastive learning alone provides the
    strongest signal for cross-modal alignment. Adding the auxiliary classification
    head ("Graph+Aux") slightly decreases MRR to 0.584 but improves mean rank performance
    from 114.1 to 93.4, indicating fewer extremely low rankings. The discriminative
    learning rate (DLR) schedule proves beneficial, as removing it ("Graph+Aux, no
    DLR") reduces MRR to 0.566.

    These ablations reveal important design considerations: (1) graph
    supervision provides the strongest boost to retrieval performance, (2) auxiliary
    classification helps regularize the model and improves worst-case performance,
    and (3) layer-specific learning rates are important for fine-tuning
    pretrained models with new architectural components. The optimal configuration
    balances these elements to achieve robust performance across different evaluation
    metrics.

    Overall, our ablation studies demonstrate that the structural contrastive
    learning approach is most effective when provided with sufficiently large
    batch sizes to capture meaningful graph context, and when complemented by appropriate
    optimization strategies to balance the different learning objectives.

    \section{Discussion}
    \subsection{Choosing the Right Hop-Distance for Positives}
    \label{sec:discussion_hops}

    \begin{figure}[h]
        \centering
        \includegraphics[width=0.9\textwidth]{
            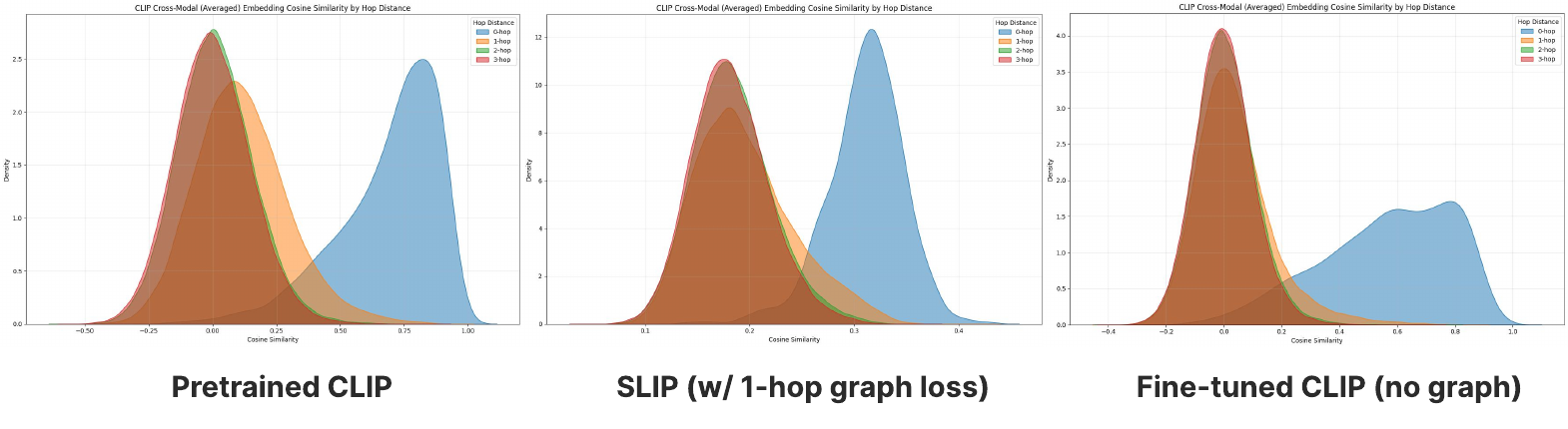
        }
        \caption{ \textbf{Cosine similarity distributions between cross-modal
        embeddings at different graph hop distances.} We show the density
        estimates of cosine similarity between image and text embeddings, grouped
        by hop distance in the product co-purchase graph: 0-hop (self), 1-hop (direct
        neighbor), 2-hop, and 3-hop.}
        \label{fig:triptych_cosine}
    \end{figure}

    A central design choice in SLIP is the selection of which $n$-hop neighbors
    to treat as positive pairs. Intuitively, immediate
    neighbors (1-hop) often represent \emph{complementary} products (e.g., a
    laptop and its charger), whereas 2, 3-hop nodes more frequently
    correspond to \emph{similar} or semantically related items (e.g., two
    different laptop models), as discussed in Sec.~\ref{sec:dataset}.

    In Fig.~\ref{fig:triptych_cosine}, we compute the cosine similarity between
    all image and text pairs within the sampled product co-purchase graph, and
    grouped by hop distances. For each training condition (e.g., pretrained CLIP,
    fine-tuned with/without graph supervision), the similarity distributions are
    then visualized using kernel density estimation (KDE) to assess how well the
    model separates semantically or structurally related nodes. Our empirical
    study of CLIP's raw cross-modal cosine-similarity distributions (on the left
    of Fig.~\ref{fig:triptych_cosine}) confirms this intuition. Before any graph-aware
    training, the 0-hop (self) similarities peak at the highest values, while 1-hop
    similarities exhibit a heavy tail extending toward the 0-hop mode,
    suggesting that some 1-hop pairs are nearly as aligned as true positives. By
    contrast, 2- and 3-hop similarities cluster tightly at lower values,
    indicating they rarely appear as strong ``false'' positives under vanilla
    CLIP.

    When we train SLIP using only 1-hop neighbors as positives (in the middle of
    Fig.~\ref{fig:triptych_cosine}), two effects emerge: (i) the 0-hop curve
    shifts leftward and sharpens, reflecting that the model has learned to distinguish
    exact matches from all others more cleanly; and (ii) the 1-hop curve shifts
    rightward, showing that structurally connected complements have been pulled closer together
    in the embedding space. This simultaneous sharpening of 0-hop and boosting
    of 1-hop similarities validate our choice of 1-hop as the \emph{sweet spot} for
    relational regularization. In contrast, a run without any graph loss (on the
    right of Fig.~\ref{fig:triptych_cosine}) fails to produce these shifts:
    the 0-hop distribution retains its heavy right tail (false positives remain),
    and the 1-hop distribution shows little movement, indicating no structural
    alignment.

    Why not use 2-hop neighbors alone? Although 2-hop items are often semantically
    similar, they also include a mix of both complementary and unrelated
    products (e.g., a laptop may be 2-hop connected to printer ink via a charger
    link). What makes things worse is that there are exponentially more 2-hop neighbors
    than 0 or 1-hop neighbors. signal. Treating all 2-hop nodes as positives risks
    introducing noise into the contrastive signal, weakening the model's ability
    to discriminate exact matches. By confining positives to 1-hop, we strike a
    balance between semantic breadth and label precision. Treating all 2-hop nodes
    as positive risks introduces noise and weakens the model's ability to discriminate exact matches. \looseness=-1

    \subsection{Towards Temporal Graph Refinement}
    Our current co-purchase graph aggregates relationships over the entire
    history of user behavior. However, consumer patterns evolve: one rarely buys
    two identical tablets on the same day, but may replace a tablet with a newer
    model years later. Consequently, static co-purchase edges can conflate \emph{complementary}
    and \emph{sequential} relationships, adding noise to both 1- and 2-hop
    neighborhoods. A promising future direction is to partition co-purchase data
    by timestamp—e.g.\ training with only edges formed within the last year—and
    to weight edges by recency. Such a \emph{time-aware} graph could yield
    cleaner 2-hop positives (truly semantically equivalent items).

    
    \bibliography{iclr2025_conference}
    \bibliographystyle{iclr2025_conference}
\end{document}